\documentclass[runningheads]{llncs}
\usepackage{graphicx}
\usepackage{tikz}
\usepackage{comment} 
\usepackage{amsmath,amssymb} % define this before the line numbering.
\usepackage{color}

\usepackage{multirow}

\usepackage{xspace}
\def\eg{\emph{e.g.}\@\xspace} 
\def\ie{\emph{i.e.}\@\xspace} 
 
\def\etc{\emph{etc}\@\xspace} 
 
\def\etal{\emph{et al}\@\xspace}

\begin{document}
\pagestyle{headings}
\mainmatter
\def\ECCVSubNumber{3160}  % Insert your submission number here

\title{RobustScanner: Dynamically Enhancing Positional Clues for Robust Text Recognition} % Replace with your title

%\begin{comment}
%\titlerunning{ECCV-20 submission ID \ECCVSubNumber} 
%\authorrunning{ECCV-20 submission ID \ECCVSubNumber} 
%\author{Anonymous ECCV submission}
%\institute{Paper ID \ECCVSubNumber}
%\end{comment}
%******************

%\begin{comment}
\titlerunning{Dynamically Enhancing Positional Clues for Robust Text Recognition}
\author{Xiaoyu Yue\inst{1}\orcidID{0000-0001-7023-0264} \and
Zhanghui Kuang\inst{1}\thanks{Zhanghui Kuang is the corresponding author.}\orcidID{0000-0002-9102-5152} \and
Chenhao Lin\inst{2} \and 
Hongbin Sun\inst{1}\orcidID{0000-0001-9183-4462} \and
Wayne Zhang\inst{1}\orcidID{0000-0002-8415-1062}}
\authorrunning{X. Yue et al.}
\institute{SenseTime Research \\ \email{\{yuexiaoyu, kuangzhanghui, sunhongbin, wayne.zhang\}@sensetime.com} \and
School of Cyber Science and Engineering, Xi'an Jiaotong University
\email{linchenhao@xjtu.edu.cn}}
%\end{comment}
%******************
\maketitle

\begin{abstract}
The attention-based encoder-decoder framework has recently achieved impressive results for scene text recognition, and many variants have emerged with improvements in recognition quality.
However, it performs poorly on contextless texts (\eg, random character sequences) which is unacceptable in most of real application scenarios. 
In this paper, we first deeply investigate the decoding process of the decoder.
We empirically find that a representative character-level sequence decoder utilizes not only context information but also positional information.
% The existing approaches heavily relying on contextual information causes the problem of attention drift.
Contextual information, which the existing approaches heavily rely on, causes the problem of attention drift.
% To suppress the side-effect of the attention drift, 
To suppress such side-effect, 
% we propose one novel position enhancement branch, 
we propose a novel position enhancement branch, 
and dynamically fuse its outputs with those of the decoder attention module for scene text recognition. 
Specifically, it contains a position aware module to
% make the encoder output 
enable the encoder to output 
feature vectors encoding their own spatial positions, and an attention module to estimate glimpses using the positional clue (\ie, the current decoding time step) only. 
The dynamic fusion is conducted for more robust feature via an element-wise gate mechanism.
Theoretically, our proposed method, dubbed \emph{RobustScanner}, decodes individual characters with dynamic ratio between context and positional clues, and utilizes more positional ones when the decoding sequences with scarce context, and thus is robust and practical.
Empirically, it has achieved new state-of-the-art results on popular regular and irregular text recognition benchmarks while without much performance drop on contextless benchmarks, validating its robustness in both 
% context
contextual
and contextless application scenarios.
\end{abstract}

\section{Introduction}
\label{sec:intro}

\begin{figure}
    \begin{minipage}{0.43\linewidth}
    \centerline{ \includegraphics[width=\linewidth]{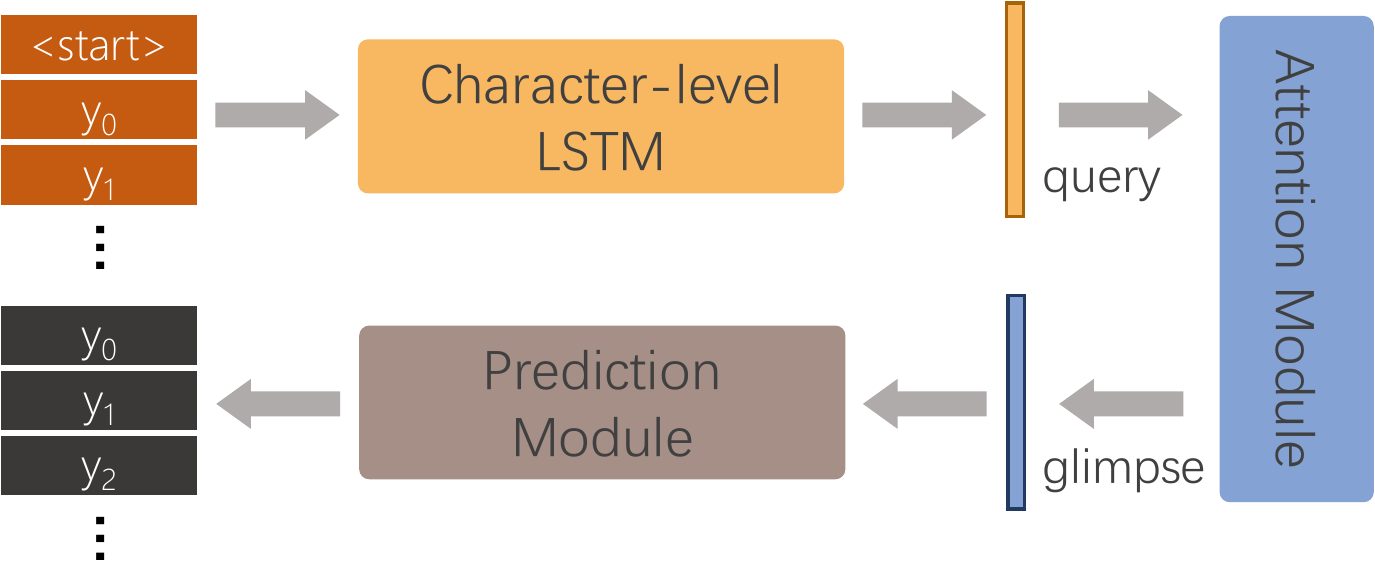}}
    \vfill
    \centerline{(a) Decoding procedure}
    \end{minipage}
    \hfill
    \begin{minipage}{0.57\linewidth}
    \centerline{ \includegraphics[width=\linewidth]{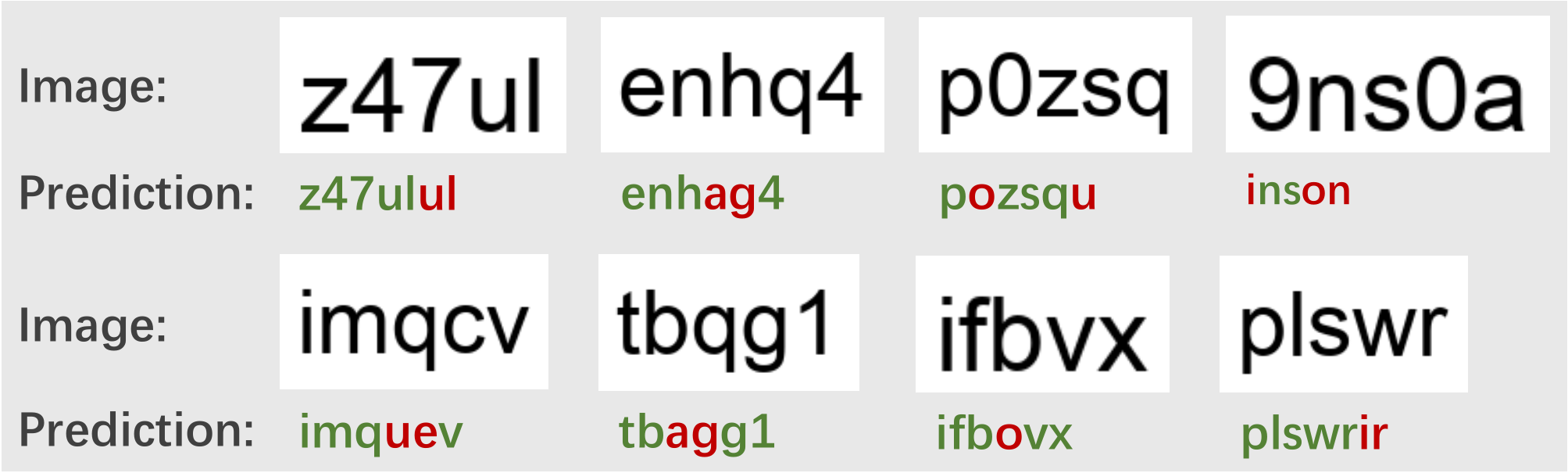}}
    \vfill
   \centerline{(b) Failure examples}
   \end{minipage}
    \caption{Illustration of the decoding procedure of encoder-decoder with attention framework and its failure cases. }
    \label{fig:overview}
\end{figure}

Scene text recognition is crucial for visual understanding and reasoning in many application scenarios~\cite{singh2019towards,biten2019scene}. Despite great progress recently, it remains a challenging task because  the potential irregularity and diversity of text shapes and layouts in the wild, which can be curved, oriented or distorted, make the misalignment between the output character sequence and the two-dimensional input image~\cite{luo2019moran,zhan2018esir,wang2019simple,li2018show,shi2018aster}.
The prevalent approaches for scene text recognition are inspired by machine translation~\cite{Sutskever,Manning2015,Bahdanau2015} and image caption~\cite{Xu2014} following the encoder-decoder framework with varied attention mechanisms~\cite{Sheng2017,wang2019simple,li2018show,yang2017learning,shi2018aster}.

Typically, the encoder-decoder with attention framework~\cite{li2018show,yang2017learning} consists of one encoder, and one decoder.
The decoder contains one character-level LSTM,
one attention module, and one prediction module.
During decoding, at each step, the LSTM takes the previously predicted character and the hidden state as inputs, and outputs one \textit{query} feature vector, which is fed into the attention module to estimate attention map and compute one \textit{glimpse} feature vector. The glimpse vector is finally classified into one character category or the $\textless\text{EOS}\textgreater$ token in the prediction module as illustrated in Figure~\ref{fig:overview} (a). 
In spite of the appealing effectiveness of the encoder-decoder with attention framework on academic benchmarks,
it performs poorly on contextless text sequence images, which hinders it from being widely used in real application scenarios.
To demonstrate it, we synthesize one benchmark of random character sequence images, dubbed \textit{RandText}, to evaluate
existing encoder-decoder with attention based methods. Surprisingly, the official released model of the state-of-the-art method SAR~\cite{li2018show} obtains a low accuracy of $59.6\%$ on it. Figure~\ref{fig:overview} (b) shows some failure samples, all of which are without background clutter, low resolutions, varied illumination or distortions, and thus considered as easy cases by human.

\begin{figure}
    \begin{minipage}{0.46\linewidth}
    \centerline{\includegraphics[width=0.7\linewidth]{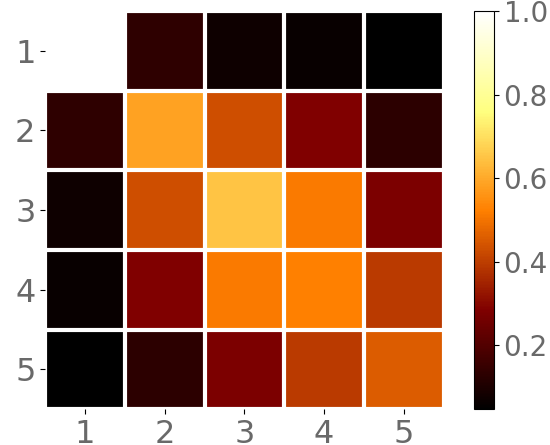}}
        \vfill
    \centerline{(a) $l=5$}    
    %\subcaption{$l=5$}
    \end{minipage}
    \hfill
    \begin{minipage}{0.46\linewidth}
    \centerline{\includegraphics[width=0.7\linewidth]{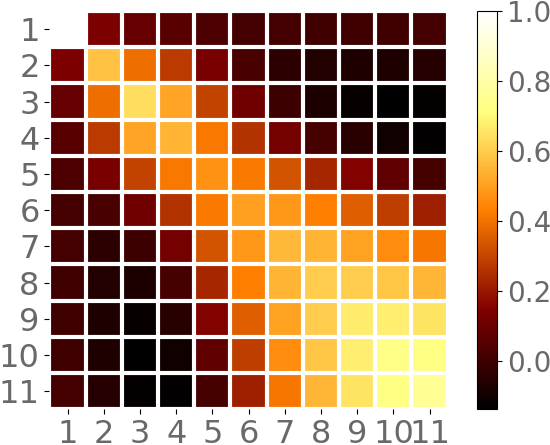}}
    %\subcaption{$l=11$}
         \vfill
    \centerline{(b) $l=11$}
    \end{minipage}
    \caption{Visualization of the averaged similarity matrix. The x-axis and the y-axis indicate the position index in sequences, while the color indicate the averaged similarity. The block at $(i,j)$ indicates $S_l(i,j)$. (a) visualizes the averaged similarity matrix when $l=5$ while (b) $l=11$.   }
    \label{fig:heatmap}
\end{figure}

To explore the underlying reasons of the failure recognition,
in this work, we deeply investigate the decoding process of encoder-decoder with attention based methods.
By computing the averaged cosine similarity between the \textit{query} feature vectors of the $i^{th}$ and $j^{th}$ time steps on ICDAR 2013 ~\cite{karatzas2013icdar} test set, we observed the high averaged similarity between those of the same time step (as shown in Figure~\ref{fig:heatmap}) although their corresponding characters are different. The observation suggests that the \textit{query} vectors encode not only context information but also the positional information. We also observed that the averaged similarity between the \textit{query} vectors of neighborhood time steps increases as the time step increases, which suggests that the positional information is drowned with the introduction of others including context information at latter time steps. It can easily lead to alignment drift and misrecognition of latter characters of contextless sequences, which coincides with most of the failure samples in Figure~\ref{fig:overview} (b).

To mitigate the above misrecognition, in this paper, we propose RobustScanner for text recognition via dynamically enhancing the positional clues of the decoder. Specifically, besides the conventional decoder, it consists of one position enhancement branch and one dynamic fusion module. The former is tailored for enhancing the conventional decoder in terms of positional encoding capability via estimating the \textit{glimpses} with positional clues (\ie, the current decoding time step) only.  The latter is designed to dynamically fuse the \textit{glimpses} of the position enhancement branch, and those of the conventional decoder via one element-wise gate mechanism. In such a way, the fusion ratio between positional and context information at each time step can be dynamically and adaptively adjusted according to the their own importance.

Our \textbf{contributions} can be summarized as follows:

\begin{itemize}
\item We investigate the intrinsic mechanism of the decoding procedure of the encoder-decoder with attention framework for the first time. We find out that the \textit{query} feature vectors of the LSTM encodes not only context but also positional information, and the context information dominates the \textit{query} at latter decoding time steps, which can lead to misrecognition on contextless text images.
\item We propose RobustScanner  to mitigate the issue of misrecogniton in contextless scenarios via  introducing a novel position enhancement branch and a dynamic fusion module.
\item We extensively evaluate our proposed RobustScanner, which achieves new state-of-the-art performances on popular regular and irregular text recognition benchmarks while without much performance drop on contextless $RandText$, validating its robustness in both context and contextless application scenarios.
\end{itemize}

\section{Related Work}

Most of traditional methods for scene text recognition~\cite{wang2011end,wang2010word,mishra2012top,lee2014region,mishra2016enhancing} adopt the bottom-up approach in which individual character is first detected by sliding window, and then integrated for taking the dependence with its neighbors into consideration. These methods might fail to detect small characters, and are easily disturbed by background clutter, illumination, and low image quality, \eg, blurring, noise, \etc. Later, top-down methods \cite{graves2006connectionist,shi2016end,he2016reading,gao2017reading,lee2016recursive,shi2016robust,bai2018edit,shi2018aster,zhan2018esir,cheng2018aon,cheng2017focusing,li2018show,wang2019simple,Qin2019,baek2019wrong,yang2019symmetry,Sun2019a,He2019} were proposed, in which text sequences are end-to-end predicted without the single character detection.
Recently, approaches which target at challenging arbitrary-shaped text recognition become dominant. These approaches can be roughly categorized into rectification-based, segmentation-based and encoder-decoder with attention-based.

\textbf{Rectification-based approaches.} They attempt to rectify irregular images to regular ones before recognition.
STN \cite{jaderberg2015spatial} was first introduced into text recognition network by Shi \etal~\cite{shi2016robust}, which was extended by considering more flexible Transformation Thin-Plate-Spline (TPS) in \cite{shi2018aster,yang2019symmetry} and conducting rectification iteratively in \cite{zhan2018esir}.
Rather than rectifying the whole text region, Liu \etal~\cite{liu2018char} detected individual characters first and then rectified them separately.
The shortness of this strategy is that, if some important information missed during rectification process, it cannot be compensated in the latter part of the network, which may be fatal at times.

\textbf{Segmentation-based approaches.} They segment each character individually to avoid the issue of irregular layout. Liao \etal \cite{liao2018scene} employed fully convolutional network to detect and recognize individual characters followed by character-combination in post-processing. 
However, they need character-level annotations which are unavailable in most public datasets. Xing \etal~\cite{Xing2019} alleviated this issue by an iterative character detection, which can transform the ability of character detection learned from synthetic data to real-world images. All the segmentation-based approaches cannot trivially extend to the text recognition with huge character dictionary, such as Chinese recognition, as they
maintain one probability heat map for each character.

\textbf{Encoder-decoder with attention-based approaches.} Most of existing state-of-the-art methods for irregular text recognition follow the encoder-decoder framework with attention mechanisms~\cite{shi2016robust,yang2017learning,cheng2017focusing,liu2018char,li2018show,yang2017learning,Qin2019,baek2019wrong,Sun2019a,wang2019simple}.
The encoder-decoder with attention framework was first proposed for NLP tasks such as machine translation ~\cite{Sutskever,Manning2015,Bahdanau2015}.
Later, it was introduced into scene text recognition to align the character in the output sequence with local image regions in~\cite{shi2016robust}.
Since then, many variants~\cite{cheng2017focusing,li2018show,yang2017learning} were emerged with improvements
in recognition quality. 
Cheng \etal~\cite{cheng2017focusing} introduced a focus network to suppress the attention drift problem.
SAR~\cite{li2018show} employed a tailored 2D attention mechanism to recognize irregular texts and achieved impressive results.
Recently, \cite{wang2019simple,Sheng2017} introduced Transformer to replace RNN structure to capture long distance context.
Wang \etal~\cite{wang2019decoupled} proposed Decoupled Attention Network (DAN) to mitigate the alignment drift problem.
Our approach falls into the encoder-decoder with attention framework.  
Similar to DAN~\cite{wang2019decoupled}, we also target at suppressing the misrecognition caused by alignment drift. Instead of decoupling attention from historical decoding results as done in~\cite{wang2019decoupled}, we propose the position enhancement branch and dynamically fusion module to adaptively adjust the ratio of positional and context clues during decoding. Besides, we deeply investigate
the intrinsic mechanism of the decoding procedure in the decoder for the first time.

Our approach is related to  the positional encoding model~\cite{vaswani2017attention,dai2019transformer,shaw2018self}. We do not target at enumerating all possible architectures and finding the optimal one of positional encoding model. In contrast, we focus on enhancing positional clues of the attention-based decoder.

\section{Methodology}
In this section, we first review the encoder-decoder with attention-based scene text recognition, and then dissect the decoder and analyze what information the \textit{query} feature vectors contain. Finally, we introduce our proposed approach RobustScanner.

\subsection{Background}
As \cite{li2018show,yang2017learning}, one representative encoder-decoder with attention-based scene text recognition approach consists of one encoder and one decoder. The decoder has one LSTM-based sequence module, one attention module and one prediction module.  Given one input image,  the encoder extracts one feature map $\mathbf{F}\in \mathbb{R}^{H\times W\times C}$. During decoding, at time step $t$, the LSTM-based sequence model first generates one hidden feature vector $\mathbf{h_t}$. Formally,

%\begin{table}[]
%\setlength{\abovecaptionskip}{10pt}
%    \centering
%    \begin{tabular}{c|c|c}
%        \hline
%        Layers & Output Size & Configurations  %\\
%        \hline
%        \multirow{2}{*}{Block 0} & $48\times 160$ & Conv, $3 \times 3$, $64$ \\
%        & $48\times 160$ & Conv, $3 \times 3$, $128$ \\
%        \hline
%        \multirow{3}{*}{Block 1} & $24\times 80$ & Max-pooling, $k$: $2\times 2$, $s$: $2\times 2$\\
%        & $24\times 80$ & $\begin{bmatrix} \text{Conv}, 3\times 3, 256 \\ \text{Conv}, 3 \times 3, 256 \end{bmatrix}$ $\times 1$ \\
%        & $24\times 80$ & Conv, $3\times 3$ , $256$ \\
%        \hline
%        \multirow{3}{*}{Block 2} & $12 \times 40$ & Max-pooling, $k$: $2\times 2$, $s$: $2\times 2$\\
%        & $12\times 40$ & $\begin{bmatrix} \text{Conv}, 3\times 3, 256 \\ \text{Conv}, 3\times 3, 256 \end{bmatrix}$ $\times 2$ \\
 %       & $12\times 40$ & \text{Conv}, $3\times 3$, $256$ \\
%        \hline
%        \multirow{3}{*}{Block 3} & $6\times 40$ & Max-pooling, $k$: $2\times 1$, $s$: $2\times 1$\\
%         & $6\times 40$ & $\begin{bmatrix} \text{Conv}, 3\times 3, 512 \\ \text{Conv}, 3\times 3, 512 \end{bmatrix}$ $\times 5$ \\
%         & $6\times 40$ & \text{Conv}, $3\times 3$, $512$ \\
%         \hline
%        \multirow{2}{*}{Block 4} & $6\times 40$ & $\begin{bmatrix} \text{Conv}, 3 \times 3, 512 \\ \text{Conv}, 3\times 3, 512 \end{bmatrix}$ $\times 3$ \\
%         & $6\times 40$ & \text{Conv}, $3 \times 3$, $512$ \\
%         \hline
%    \end{tabular}
%    \caption{The architecture of the backbone.}
%    \label{tab:cnn_architecture}
%\end{table}

\begin{equation}
    \mathbf{h_t} = \text{LSTM}(\mathbf{x_t}, \mathbf{h_{t-1}}), \mathbf{x_t} = \begin{cases} \mathbf{y_{t-1}} & \text{if } t>1 \\ \textless\text{start}\textgreater & \text{if } t=1\end{cases},
    \label{eq:lstm}
\end{equation}
where $\textless\text{start}\textgreater$ is a special start token, and $\mathbf{y_{t-1}}$ is the output of decoding process at time step $t-1$. $\mathbf{h_t}$ is considered as
the \textit{query} feature vector of the attention module to compute one attention map as follows:
\begin{equation}
    \alpha_{ij}^t = \text{softmax}(\mathbf{h_t}^{T}\mathbf{f_{i,j}}),
    \label{eq:attention}
\end{equation}
where $\mathbf{h_t}^{T}$ indicates the transpose of the vector $\mathbf{h_t}$, and $\mathbf{f_{i,j}}$ represents the feature vector at the position $(i, j)$ of $\mathbf{F}$. The \textit{glimpse} vector $\mathbf{g_t}$ is then computed by the weighted aggregation of the convolutional feature map $\mathbf{F}$ as follows:
\begin{equation}
    \mathbf{g_t} = \sum_{ij} \alpha_{ij}^t\mathbf{f_{i,j}}.
    \label{eq:glimpse}
\end{equation}
Finally, the glimpse vector $\mathbf{g_t}$ is classified into one character or the $\textless\text{EOS}\textgreater$ token.
Formally,
\begin{equation}
    \mathbf{y_t} = \text{softmax}(\mathbf{W}\mathbf{g_t} + b),
    \label{eq:classifier}
\end{equation}
where $\mathbf{W}$ and $b$ indicate the linear transformation and the bias of the classifier respectively.
\subsection{Decoder Dissection}
\label{sec:sec_decoder_dissection}
From Equation (\ref{eq:lstm}), (\ref{eq:attention}), (\ref{eq:glimpse}), and (\ref{eq:classifier}), we conclude that given the feature map $\mathbf{F}$ and learned network parameters, the recognized character $\mathbf{y_t}$ depends on the \textit{query} vector $\mathbf{h_t}$ only. What information does the \textit{query} vector $\mathbf{h_t}$ encode so that its corresponding attention weight $\alpha^t$ can highlight and the classifier can correctly recognize the $t^{th}$ character on the input image?  Obviously, the \textit{query} vector contains context information since it depends on previously-predicted characters from Equation (\ref{eq:lstm}).

We observe that the \textit{query} vector $\mathbf{h_1}$ keeps unchanged for different text sequences, and  does not encode any context information.
However, the first character can still be correctly recognized. This implies that the \textit{query} vector $\mathbf{h_t}$ (including $\mathbf{h_1}$) contains the positional information. \ie, the character index in one sequence.

To verify the above conjecture, we analyze the similarity of the \textit{query} for different text sequences at the same time step.
We conduct experiments on test set of ICDAR 2013~\cite{karatzas2013icdar}.
Let $\mathcal{I}$ denote the set of all text sequences, and $\widehat{\mathcal{I}_l}$ be the subset of $\mathcal{I}$ with all text sequences of length $l$.
We compute the average cosine similarity $S_l(i, j)$ between the \textit{query} vectors at position $i$ and $j$ over $\widehat{\mathcal{I}_l}$ as follows:
\begin{equation}
    S_l(i, j) = \frac{\sum_{m\neq n}^{\left | \widehat{\mathcal{I}_l} \right |} \cos (\mathbf{h_{i}^{m,l}}, \mathbf{h_{j}^{n,l}})}{\left | \widehat{\mathcal{I}_l} \right |({\left | \widehat{\mathcal{I}_l} \right |}-1)},
\end{equation}
where $\left | \widehat{\mathcal{I}_l} \right |$ denotes the cardinality of $\mathcal{I}_l$, and $\mathbf{h_{i}^{m}}$ denotes the $i^{th}$ \textit{query} vector for the $m^{th}$ sequence in $\mathcal{I}_l$.

Figure~\ref{fig:heatmap} visualizes the averaged similarity matrixes when $l=5$ and $l=11$. We have two observations:
\begin{itemize}
\item Averaged similarities between the \textit{query} vectors of the same position are obviously
higher than those of different positions.
\item As the time step increases, the contrast between the averaged similarities between the \textit{query} vectors of the same position, and those of neighbor positions become smaller.
\end{itemize}
Note that the $i^{th}$ character of sequences in $\mathcal{I}_l$ varies from one sequence to another.
High value of $S_l(i,i)$ suggests queries can be well separated into groups corresponding to their steps using kernel methods. We further conduct the linear regression between queries ($\mathbf{h_t}$) and positions ($t$) via fitting $t = \mathbf{W_r}\mathbf{h_t}+b_r$ on the data with $l=5$ (90\% for training, 10\% for test). We report the classic R-squared $R^2$ which indicates the proportion of the variance of the dependent variable that is predictable. We get $R^2=0.994$ and $R^2=0.956$ on training and test set. High $R^2$ scores prove queries can reliably predict positions. Thus, queries encode them. The second observation suggests that the positional information is drowned with the introduction of others including the context information. 

\subsection{RobustScanner}  
\label{sec:d2fnet}
\begin{figure*}
    \centering
    \begin{tabular}{c}
        \includegraphics[width=\linewidth]{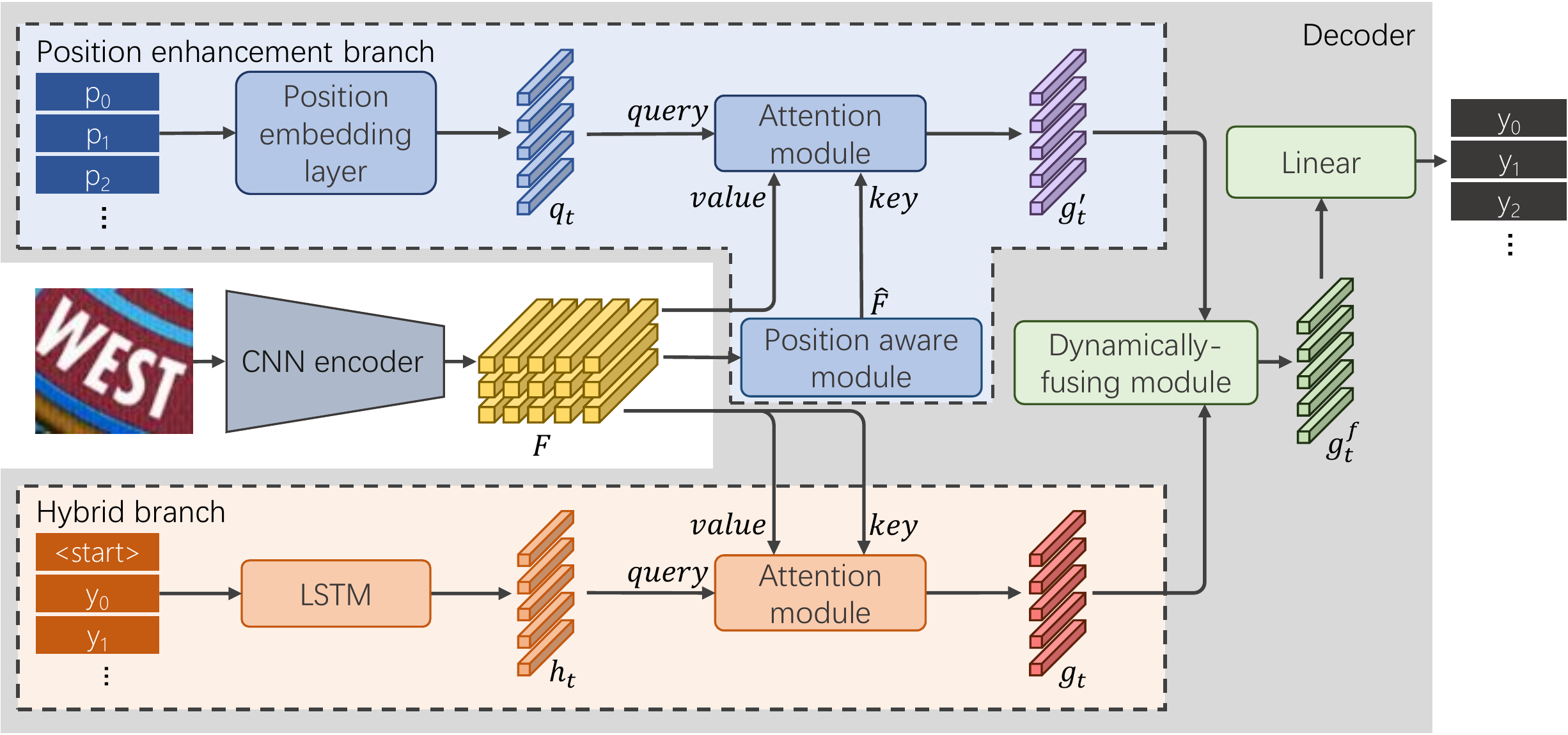}
    \end{tabular}
    \caption{Architecture of RobustScanner.  Given an input image, we first extract its feature map $\mathbf{F}$ by the CNN encoder. $\mathbf{F}$ is then fed into the hybrid branch, and the position enhancement branch, obtaining the \textit{glimpses} $\mathbf{g_t}$ and $\mathbf{g'_t}$, which are dynamically fused by the dynamically-fusing module before predicting the $t^{th}$ character. }
    \label{fig:framework}
\end{figure*}

\textbf{Overview.} As discussed in Section~\ref{sec:sec_decoder_dissection}, the \textit{query} vectors in the encoder-decoder with attention based framework contain the hybrid information of context and position. The positional clues become weaker while the contextual ones become stronger as the time step increases during decoding, which may lead to alignment drift and misrecognition, especially on contextless text images.
To enhance the positional information, we propose one position enhancement branch and dynamically fuse its outputs with those of the conventional decoder.
As shown in Figure~\ref{fig:framework}, our RobustScanner consists of one encoder and one decoder. In the encoder, we adapt one 31-layer ResNet~\cite{he2016deep} as backbone as done in \cite{li2018show}. %The backbone downsamples input images by 8$\times$ and 4$\times$ along vertical and horizontal directions respectively. Its detailed architecture is presented in Table~\ref{tab:cnn_architecture}.
The decoder consists of one hybrid branch,  one position enhancement branch, one dynamically-fusing module, and one prediction module.

\textbf{Hybrid Branch.}
The hybrid branch consists of one two-layer LSTM with 128 hidden state size and one attention module. 
The LSTM takes the previously-predicted character and its hidden variable as input and generates the \textit{query} vector $\mathbf{h_t}$. Then the \textit{query} vector is fed into the attention module to estimate \textit{glimpse} vector $\mathbf{g_t}$ for the character prediction during decoding (see Equation (\ref{eq:lstm}) and (\ref{eq:glimpse})). It utilizes both contextual and positional information simultaneously.

\begin{figure}
    \begin{minipage}{0.5\linewidth}
    \centerline{\includegraphics[height=1.1\linewidth]{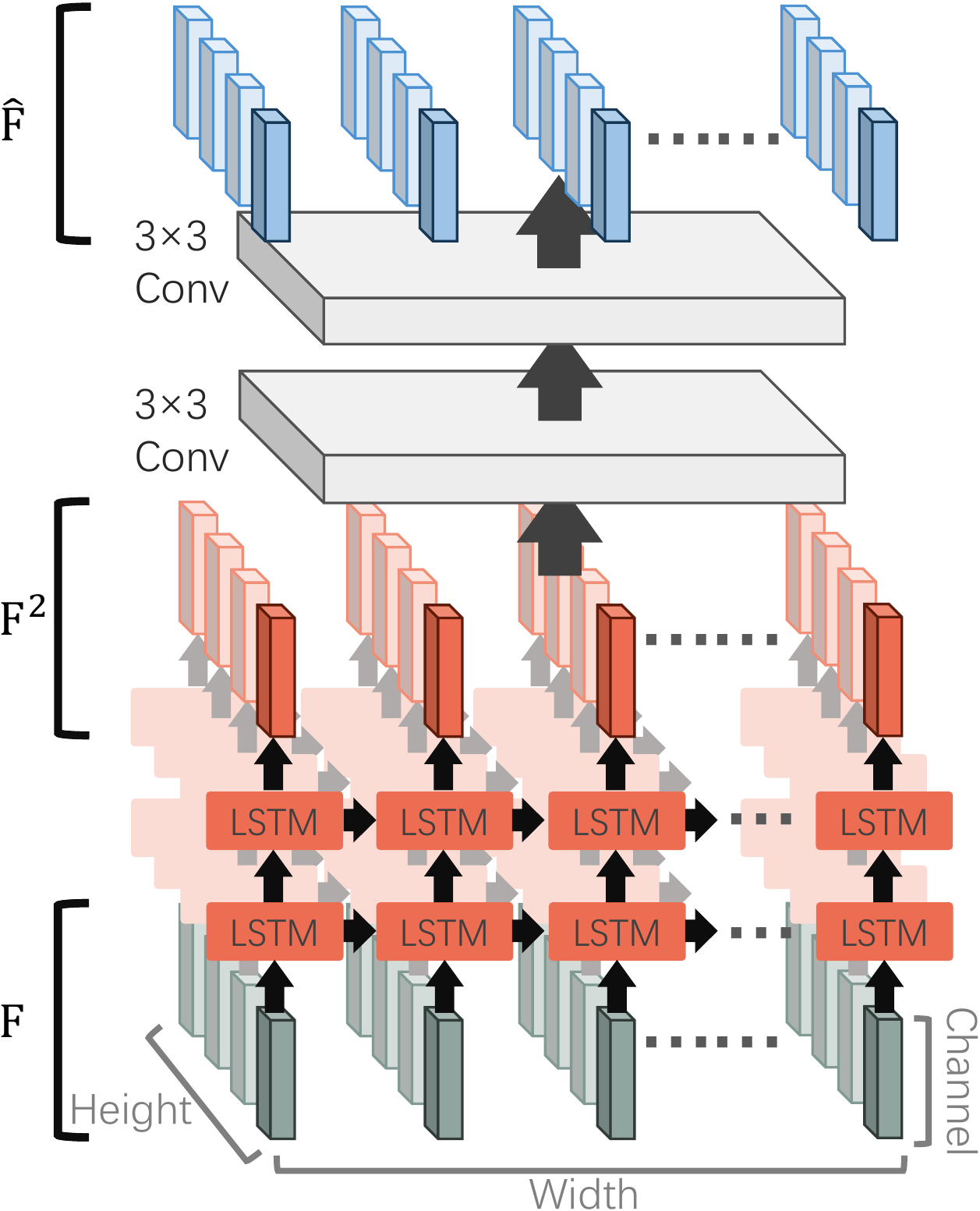}}
        \vfill
    \centerline{(a) Position aware module}
    %\subcaption{$l=5$}
    \end{minipage}
    \hfill
    \begin{minipage}{0.5\linewidth}
    \centerline{\includegraphics[height=1.1\linewidth]{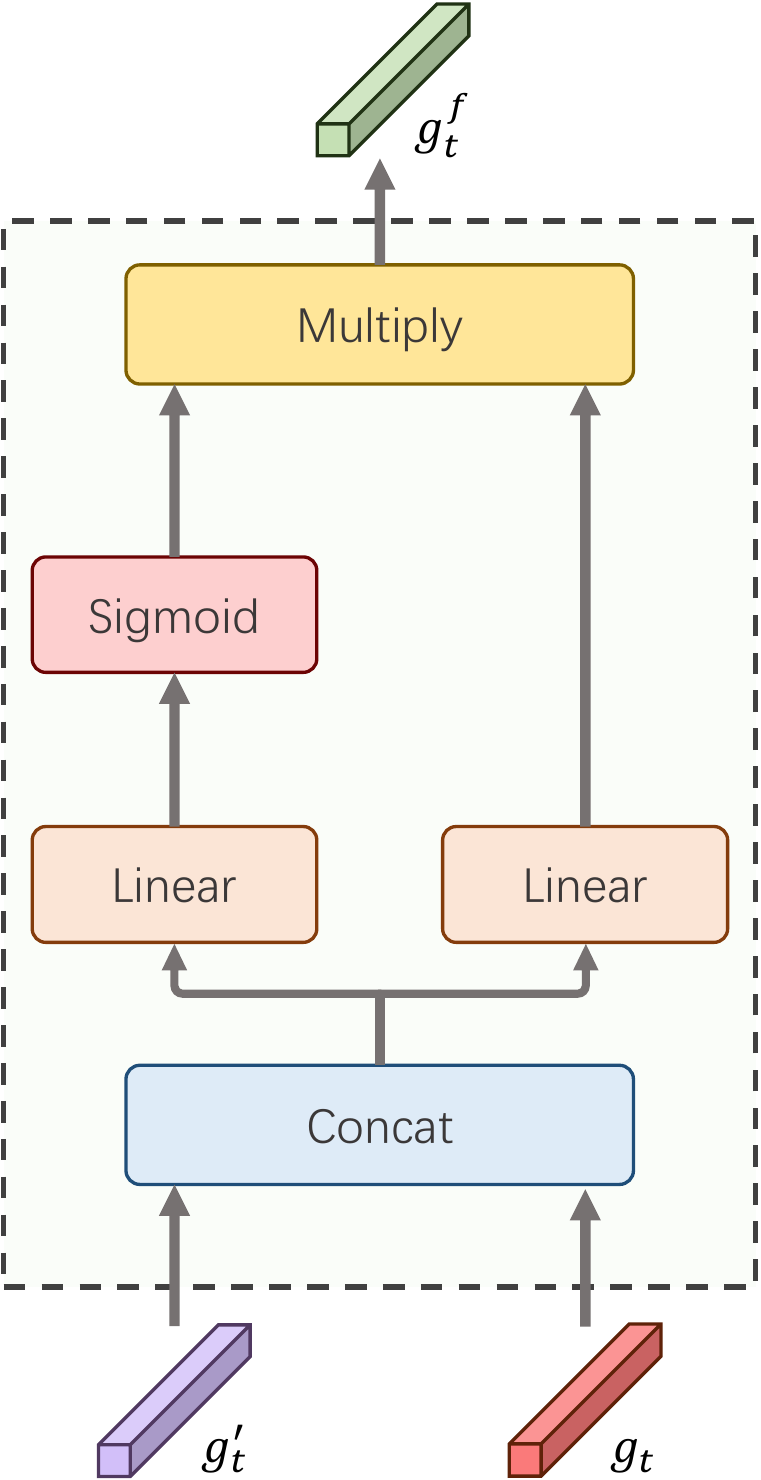}}
    %\subcaption{$l=11$}
         \vfill
    \centerline{(b) Dynamically-fusing module}
    \end{minipage}
    \caption{Architecture of the position aware module and that of the dynamically-fusing module. (a) the position aware module inputs $\mathbf{F}$, then obtains $\mathbf{F^2}$ after two LSTM layers, and finally outputs $\mathbf{\widehat{F}}$ after two $3\times3$ convolutional layers. (b)the dynamically-fusing module takes $\mathbf{g_t}$ and $\mathbf{g'_t}$ as inputs and output one adaptively fused glimpse $\mathbf{g_t^f}$. }
    \label{fig:position_aware}
\end{figure}

\textbf{Position Enhancement Branch.}
%The positional information becomes weak while the context information becomes strong in the latter time steps when decoding.
The positional information becomes weak while the contextual information becomes strong in the latter time steps during decoding. It would lead to the alignment drift and serious misrecongition on contextless text images where context cannot be used to reliably predict characters.
Our position enhancement branch is designed to mitigate this problem. It consists one position embedding layer, one position aware module and one attention module.

The position embedding layer encodes the decoding time step. It inputs one hot vectors and outputs one specific embedding vector $\mathbf{q_t}$ for each time step $t$. Note that the embedding vector of each time step keeps unchanged across different decoding sequences.

\textit{Position Aware Module.}
%Although the position embedding layer encodes the character index in sequences, precisely locating the $t^{th}$ character on the encoder output feature map $\mathbf{F}$ is challenging. It requires the global information. 
Although the position embedding layer encodes the character index in sequences, precisely locating the $t^{th}$ character on the encoder output feature map $\mathbf{F}$, which requires the global information, is still challenging.
However, the width of input image might be beyond the receptive field size of the CNN encoder. Moreover, counting the indexes of characters in images needs to understand the shapes and directions of the text sequences. To this end, we propose a position aware module to capture the global and high-level information so that the encoder output feature map is position-aware. 
Figure~\ref{fig:position_aware} (a) shows the detailed architecture of the proposed position aware module. We employ two-layer LSTM with 128 hidden state size for each row of the feature map $\mathbf{F}$ to capture the global context. For all rows, the LSTMs share parameters to overcome overfitting and reduce parameter amount. We then employ two $3\times 3$ convolutional layers with one ReLU between them to generate the \textit{key} vectors $\mathbf{\widehat F}$ of its following attention module.
  Formally,
%\begin{equation}
\begin{align}
    \mathbf{f_{i,j}^{1}} &= \text{LSTM}(\mathbf{f_{i,j}}, \mathbf{f_{i,j-1}^{1}}),\\
    \mathbf{f_{i,j}^2} &= \text{LSTM}(\mathbf{f_{i,j}^{1}}, \mathbf{f_{i,j-1}^2}), \\
    \mathbf{\widehat F} &=f(\mathbf{F^2}),
\end{align}
where $\mathbf{f_{i,j}^{1}}$ and $\mathbf{f_{i,j}^{2}}$ are the first and the second LSTM hidden variables respectively. $\mathbf{F^2}$ is the tensor consisting of $\mathbf{f_{i,j}^{2}}$, and $f(\cdot)$ is the stack of $3\times 3$ \textit{conv}, ReLU, and $3\times 3$ \textit{conv} operations.

We feed the output vectors $\mathbf{q_t}$ of the embedding layer into the attention module as \textit{key} vectors to estimate attention maps, which are used to compute the final \textit{glimpse} vectors $\mathbf{g'_t}$ as done in Equation (\ref{eq:attention}) and (\ref{eq:glimpse}). Different from conventional methods, we use the position aware feature map $\mathbf{\widehat F}$ instead of $\mathbf{F}$ when computing the attention maps. 

Note that position aware module injects the global context so that it has the capability to output position aware features. The query vectors in the position enhancement branch explicitly encode the character order indexes (positions). The end-to-end training enforces  the position aware module to output feature maps correlated with positions so that the attention module in the branch can correctly output the feature glimpse at each decoding step (position) with the position aware module output feature maps and position-specific query vector as inputs (see Figure~\ref{fig:framework}). Otherwise, texts cannot be correctly decoded. 

\textbf{Dynamically-Fusing Module.}
As shown in Figure~\ref{fig:position_aware} (b), we propose to dynamically fuse the hybrid branch output $\mathbf{g_t}$ and the position enhancement branch output $\mathbf{g'_t}$
at each time step $t$. We design one gate mechanism to predict one attention weight 
for each dimension of their concatenation, which is used to enhance or suppress their corresponding feature.  Formally,

\begin{align}
    \mathbf{w_t} &= \text{sigmoid}(\mathbf{W_a}[\mathbf{g_t}; \mathbf{g'_t}]),\\
    \mathbf{g_t^{f}} &= \mathbf{w_t}\odot(\mathbf{W_p}[\mathbf{g_t}; \mathbf{g'_t}]),
\end{align}
where $\mathbf{W_a}$ and $\mathbf{W_p}$ are two learned linear transformations. $\odot$ indicates the element-wise multiplication operation.
$\mathbf{g_t^{f}}$ is the final output of our dynamically-fusion module, which is used to predict the character via the prediction module.
\section{Experiments}
%In this section, we extensively conduct experiments to validate the effectiveness of the proposed RobustScanner. We first introduce the academical datasets and our RandText dataset. Then, we describe the implementation details. 
%We finally compare our proposed RobustScanner with state-of-the-art methods for scene text recognition.
%In addition, ablation studies are performed to evaluate the effectiveness of each component of our RobustScanner. 

\subsection{Datasets}
\label{sec:dataset}

For fair comparison with previous state-of-the-art approaches, we follow their settings to train the proposed network by using two public available synthetic datasets, \ie MJSynth~\cite{jaderberg2014synthetic} and SynthText~\cite{gupta2016synthetic}.
We conduct extensively experiments on 6 standard benchmarks including 3 regular text datasets (IIIT 5K-words~\cite{mishra2012scene}, Street View Text~\cite{wang2011end}, ICDAR 2013~\cite{karatzas2013icdar}), 3 irregular text datasets (ICDAR 2015~\cite{karatzas2015icdar}, Street View Text Perspective~\cite{quy2013recognizing}, CUTE 80~\cite{risnumawan2014robust}). 

\textbf{IIIT 5K-words} (IIIT5K)~\cite{mishra2012scene} is a large dataset containing $5000$ word patches cropped from natural scene images collected from Google image search, in which $2000$ images are used for training and $3000$ for test. Text instances in these patches are regular with the horizontal layout.

\textbf{Street View Text} (SVT)~\cite{wang2011end} consists of $647$ word patches cropped from Google Street View for testing. Most images are horizontal, but severely corrupted by noise and blur, or with low-resolution.

\textbf{ICDAR 2013}~\cite{karatzas2013icdar} has $848$ cropped word patches for training and $1095$ for test. For fair comparison with other reported results, we discard images that contain non-alphanumeric characters, which results in $1015$ test patches. Words in this dataset are mostly regular.

\textbf{ICDAR 2015}~\cite{karatzas2015icdar} contains word patches cropped from incidental scene images captured under arbitrary angles. Hence most word patches in this dataset are irregular (oriented, perspective or curved). It contains $4468$ patches for training and $2077$ for test.

\textbf{Street View Text Perspective} (SVTP)~\cite{quy2013recognizing} consists of $639$ word patches, which are cropped from side view snapshots in Google Street View and encounter severe perspective distortions. All patches are used for test.

\textbf{CUTE 80}~\cite{risnumawan2014robust} contains 288 cropped high resolution images for test and many of them are curved irregular text images.

\textbf{RandText} contains $500$ test images. We synthesize them by pasting black random character sequences on white background images. The character set contains small and capital letters, and numbers. Some examples can be found in Figure~\ref{fig:overview} (b). We will publicly release RandText to facilitate the future research.

\subsection{Implementation Details}

The proposed framework is implemented by using PyTorch. All experiments are conducted on servers with 4 NVIDIA Titan X GPUs with 12 GB memory.

Our RobustScanner is trained from scratch using Adam optimizer~\cite{kingma2014adam} with the base learning rate $1\times 10^{-3}$. The whole training process contains $5$ epochs, while the learning rate decreases to $1\times 10^{-4}$ at the $3^{rd}$ epoch and $1\times 10^{-5}$ at the $4^{th}$ epoch.
The batch size at training phase is set to $128$.
For both training and test, heights of all image patches are set to 48, while widths are proportionally scaled with heights, but no longer than 160 and no smaller than 48 pixels.
Our models recognizes 91 token classes, including 10 digits, 52 case sensitive letters, 28 punctuation characters and an $\textless\text{EOS}\textgreater$ token.

To reduce the computation time of two attention module, the dimension of the hybrid query vector $h_t$ and the position embedding $q_t$ is set to 128. Before feeding $F$ to the hybrid attention module and the position aware module, a $1\times1$ \textit{conv} layer is added to reduce the channel dimension to 128.
The maximum number of position embeddings is set to 36.

In the test stage, for images with height larger than width, we will rotate the image by $90$ degree clockwise and anticlockwise respectively, and recognize them together with the original image. For the recognition results of each 3-sibling image group, the top-score one will be chosen as the final recognition result.

\subsection{Comparison with State-of-the-art Approaches}

\begin{table*}[]
\setlength{\abovecaptionskip}{10pt}
\begin{minipage}{\linewidth}
\resizebox{\columnwidth}{!}{
\centering
    \begin{tabular}{c|c|c|c|c|c|c|c}
    \hline
    \multirow{2}{*}{Method} & \multirow{2}{*}{Training Data} & \multicolumn{3}{c|}{Regular Text} & \multicolumn{3}{c}{Irregular Text} \\
     \cline{3-8}
  & & IIIT5K & SVT & ICDAR 2013 & ICDAR 2015 & SVTP & CUTE 80 \\
\hline
% Liu \etal~\cite{liu2018squeezedtext} & - & 87.0 & - & 92.9 & - & - & - \\
% Shi \etal~\cite{shi2016end} & MJ & 78.2 & 80.8 & 86.7 & - & - & - \\
% Shi \etal~\cite{shi2016robust} & MJ & 81.9 & 81.9 & 87.5 & - & - & - \\
% Lee and Osindero~\cite{lee2016recursive} & MJ & 78.4 & 80.7 & 90.0 & - & - & - \\
% Gao \etal~\cite{gao2017reading} & MJ & 81.8 & 82.7 & 88.0 & - & - & - \\
% Liu \etal~\cite{liu2018char} & MJ & 83.6 & 84.4 & 90.8 & - & 73.5 & - \\
% Wu \etal~\cite{wu2018scan} & MJ & 84.2 & 85.0 & 90.4 & - & - & - \\
Cheng \etal~\cite{cheng2017focusing} & MJ + ST & 87.4 & 85.9 & 93.3 & 70.6 & - & - \\
Cheng \etal~\cite{cheng2018aon} & MJ + ST & 87.0 & 82.8 & - & 68.2 & 73.0 & 76.8 \\
Shi \etal~\cite{shi2018aster} & MJ + ST & 93.4 & \textbf{93.6} & 91.8 & 76.1 & 78.5 & 79.5 \\
Zhan and Lu \cite{zhan2018esir} & MJ + ST & 93.3 & 90.2 & 91.3 & 76.9 & 79.6 & 83.3 \\
Gao \etal~\cite{gao2018recurrent} & MJ + ST & 94.0 & 88.6 & 93.2 & \textbf{77.1} & 80.6 & 88.5\\
Bai \etal~\cite{bai2018edit} & MJ + ST & 88.3 & 87.5 & 94.4 & 73.9 & - & - \\
Luo \etal~\cite{luo2019moran} & MJ + ST & 91.2 & 88.3 & 92.4 & 68.8 & 76.1 & 77.4 \\
Wang \etal~\cite{wang2019simple} & MJ + ST & 93.3 & 88.1 & 91.3 & 74.0 & 80.2 & 85.1 \\
Lyu \etal~\cite{lyu20192d} & MJ + ST & 94.0 & 90.1 & 92.7 & 76.3 & 82.3 & 86.8 \\
% Yang \etal~\cite{yang2019symmetry} & MJ + ST & 94.4 & 88.9 & 93.9 & 78.7 & 80.8 & 87.5 \\
Xie \etal~\cite{xie2019aggregation} & MJ + ST & 82.3 & 82.6 & 89.7 & 68.9 & 70.1 & 82.6 \\
DAN~\cite{wang2019decoupled} & MJ + ST & 94.3 & 89.2 & 93.9 & 74.5 & 80.0 & 84.4 \\
Bartz \etal~\cite{bartz2019kiss} & MJ + ST & 94.6 & 89.2 & 93.1 & 74.2 & \textbf{83.1} & 89.6 \\
Bleeker \etal~\cite{bleeker2019bidirectional} & MJ + ST & 94.7 & 89.0 & 93.4 & 75.7 & 80.6 & 82.5 \\
Long \etal~\cite{long2020new} & MJ + ST & 93.7 & 88.9 & 92.4 & 76.6 & 78.8 & 86.8 \\
Baek \etal~\cite{baek2019wrong} & MJ + ST & 87.9 & 87.5 & 92.3 & 71.8 & 79.2 & 74.0 \\
% Wan \etal~\cite{wan2019textscanner} & MJ + ST & 93.9 & 90.1 & 92.9 & 79.4 & 84.3 & 83.3 \\
\hline
RobustScanner & MJ + ST & \textbf{95.3} & 88.1 & \textbf{94.8} & \textbf{77.1} & 79.5 & \textbf{90.3} \\
\hline
\hline
SAR~\cite{li2018show} & MJ + ST + R & 95.0 & \textbf{91.2} & 94.0 & 78.8 & \textbf{86.4} & 89.6 \\
\hline
RobustScanner & MJ + ST + R & \textbf{95.4} & 89.3 & \textbf{94.1} & \textbf{79.2} & 82.9 & \textbf{92.4} \\
\hline
% \hline
% Yang \etal~\cite{yang2017learning} & PRI & - & - & - & - & 75.8 & 69.3 \\
% Liao \etal~\cite{liao2018scene} & ST + C & 91.9 & 86.4 & 91.5 & - & - & 79.9 \\
% Liu \etal~\cite{liu2018synthetically} & - & 89.4 & 87.1 & 94.0 & - & 73.9 & 62.5 \\
    \end{tabular}
}
\end{minipage}
    \caption{Comparison with state-of-the-art methods. ``MJ'', ``ST'' and ``R'' are the training data of MJSynth~\cite{jaderberg2014synthetic}, SynthText~\cite{gupta2016synthetic} and training splits of real datasets, respectively.}
    \label{tab:state_of_the_arts}
\end{table*}

We compare our approach with previous state-of-the-art methods in Table~\ref{tab:state_of_the_arts}.  It has been shown that our proposed RobustScanner achieves best results on four datasets including IIIT5K, ICDAR 2013, ICDAR 2015 and CUTE 80 when only
synthetic training datasets are used. To demonstrate the potential of our proposed method, we further train it on MJSynth, SynthText, and training sets of real datasets in Section~\ref{sec:dataset} following the training setting of SAR~\cite{li2018show}. We have observed that RobustScanner outperforms its competitor SAR on four out of six benchmarks. Especially, it obtains the accuracy of 92.4\%, and outperforms SAR with impressive margins on the challenging irregular text dataset CUTE 80. It can be noticed that our method performs worse than SAR~\cite{li2018show} on two street view text datasets, we attribute the performance gap to that SVT and SVTP consist of contextual words and suffer from low resolution, blur and distortion, where context is more important than positions when recognizing texts.

%\begin{table*}{}
%\centering
%\begin{tabular}{c|c|c}
%\hline
%Method & Training Data & RandText \\
%\hline
%SAR~\cite{li2018show} & MJ + ST & 59.6 \\
%SAR~\cite{li2018show} & MJ + ST + R & 67 \\
%\hline
%DAN~\cite{wang2019decoupled} & MJ + ST & 76.4 \\
%\hline
%Wang \etal~\cite{wang2019simple} & MJ + ST & %78.8 \\
%\hline
%RobustScanner & MJ + ST & 81.2 \\
%\hline
%\end{tabular}
\begin{table*}{}
\setlength{\abovecaptionskip}{10pt}
\centering
\begin{tabular}{c|c|c|c|c}
\hline
Method & SAR~\cite{li2018show} & DAN~\cite{wang2019decoupled}& Wang \etal~\cite{wang2019simple}&RobustScanner\\ \hline
Accuracy&59.6&76.4&78.8&\textbf{81.2}\\ \hline
%Method & Training Data & RandText \\
%\hline
%SAR~\cite{li2018show} & MJ + ST & 59.6 \\
%SAR~\cite{li2018show} & MJ + ST + R & 67 \\
%\hline
%DAN~\cite{wang2019decoupled} & MJ + ST & 76.4 \\
%\hline
%Wang \etal~\cite{wang2019simple} & MJ + ST & %78.8 \\
%\hline
%RobustScanner & MJ + ST & 81.2 \\
%s\hline
\end{tabular}

\caption{Comparison on RandText. All methods are trained on the training data of MJSynth~\cite{jaderberg2014synthetic} and SynthText~\cite{gupta2016synthetic} for fair comparison.}
\label{tab:randtext}
\end{table*}

This paper targets at mitigating the misrecongition issue on contextless text images of the encoder-decoder with attention based framework. Therefore, we evaluate our proposed 
RobustScanner on RandText dataset. Table~\ref{tab:randtext} compared it with representative encoder-decoder with attention based approaches. Again, our proposed RobustScanner obviously outperforms its counterparts including LSTM based methods~\cite{li2018show} and ~\cite{wang2019decoupled} and the Transformer based method~\cite{wang2019simple}. Especially,
DAN~\cite{wang2019decoupled} decouples the historical decoded characters from the attention estimation to suppress the alignment drift problem. However, it still performs worse than our RobustScanner by absolute 4.8\%.  The effectiveness of RobustScanner on both context and contextless benchmarks validate its robustness in real application scenarios.

\subsection{Ablation Study}
\textbf{The effectiveness of each branch.}
To show the effectiveness of each branch, we evaluate our RobustScanner without the position enhancement branch or the hybrid branch in Table~\ref{tab:branch}. In these experiments, the \textit{glimpse} vectors are fed directly into the classifier to decode the character at each time step without dynamic fusion. We have two observations. First, RobustScanner greatly outperforms its counterpart without the hybrid branch on all academical regular and irregular benchmarks, and performs slightly worse on RandText. Due to the collection bias, all academical benchmarks have abundant context. It suggests that the context clues introduced via the hybrid branch can boost the recognition performances on context text images, and causes side effect on contextless text images.   Second, the performance of RobustScanner drops drastically when without applying the position enhancement branch on both context and contextless text images. 
% Particularly, RobustScanner without the hybrid branch obtains a low accuracy of 46.8\% on RandText, which is lower than it with both branches by absolutely 34.4\%.
Particularly, RobustScanner without the position enhancement branch obtains a low accuracy of 46.8\% on RandText, which is lower than it with both branches by absolutely 34.4\%.
We conclude that the capacity of encoding positional clues in the decoder with attention should be enhanced, especially when recognizing contextless text images. 

\begin{table*}{}
\setlength{\abovecaptionskip}{10pt}
\begin{minipage}{\linewidth}
\resizebox{\columnwidth}{!}{
\centering
\begin{tabular}{c|c|c|c|c|c|c|c}
\hline
 & IIIT5K & SVT & ICDAR 2013 & ICDAR 2015 & SVTP & CUTE 80 & RandText \\
\hline
w/o HB & 92.5 & 84.4 & 91.4 & 70.7 & 74.4 & 83.3 & \textbf{83.4}\\
w/o PEB & 91.0 & \textbf{88.3} & 91.7 & 72.0 & 75.4 & 84.0 & 46.8\\\hline
RobustScanner&\textbf{95.3}&88.1&\textbf{94.8}&\textbf{77.1}&\textbf{79.5}&\textbf{90.3}&81.2 \\
\hline
\end{tabular}
}
\end{minipage}
\caption{Evaluation of the effectiveness of each branch in RobustScanner. HB and PEB indicate the hybrid branch and the position enhancement branch respectively.}
\label{tab:branch}
\end{table*}

\textbf{The effectiveness of the position aware module.} Enhancing position encoding is nontrivial since it requires the global context of the input image and the comprehension of the shapes and directions of text sequences. Our proposed position aware module is the core component of the position enhancement branch. Table~\ref{tab:pam}
evaluates its effect. if we take out the position aware module, our RobustScanner decreases from 92.5\%, 84.4\%, 91.4\%, 70.7\%, 74.4\%, 83.3\% and 83.4\% to 88.7\%, 84.1\%, 89.0\%, 61.8\%, 62.3\%, 76.7\% and 68.0\% on IIIT5K, SVT, ICDAR 2013, ICDAR 2015, SVTP, CUTE 80 and RandText respectively.  We also compare our position aware module with sine and cosine positional encoding~\cite{vaswani2017attention}. We compute the 128-dimensional sine and cosine positional encoding via $pe(pos,2i)=\sin(pos/1000^{2i/128})$ and
$pe(pos,2i+1)=\cos(pos/1000^{2i/128})$ with $pos$ being the horizontal coordinate of the current encoder output feature vector in the spatial space, and $2i$ and $2i+1$ being the encoding indexes. We pad the sine and cosine positional encoding with $\mathbf{F}$ and obtain $\mathbf{\widehat F}$, which is fed into the attention module as \textit{key} vectors. As shown in Table~\ref{tab:pam}, sine and cosine positional encoding leads to obvious performance drop. Arguably, it is because since and cosine positional encoding contain absolute coordinate information only while relative  character indexes in sequences are required in the position enhancement branch. 
\begin{table*}{}
\setlength{\abovecaptionskip}{10pt}
\begin{minipage}{\linewidth}
\centering
\begin{tabular}{c|c|c|c|c|c|c|c}
\hline
 & IIIT5K & SVT & ICDAR 2013 & ICDAR 2015 & SVTP & CUTE 80 & RandText \\
\hline
w/o PAM & 88.7 & 84.1 & 89.0 & 61.8 & 62.3 & 76.7 & 68.0 \\
sine\&cosine & 86.1 & 83.2 & 87.8 & 57.5 & 56.1 & 74.0 & 49.0 \\
RobustScanner & \textbf{92.5} & \textbf{84.4} & \textbf{91.4} & \textbf{70.7} & \textbf{74.4} & \textbf{83.3} & \textbf{83.4} \\
\hline
\end{tabular}
\end{minipage}
\caption{Evaluation of the effectiveness of the position aware module, and other position encoding method. PAM indicates the position aware module. sine\&cosine indicates padding sine and cosine positional encoding with the encoder output feature maps.}
\label{tab:pam}
\end{table*}

\textbf{The effectiveness of the dynamically-fusing module.}  Table~\ref{tab:df} compares our proposed dynamically-fusing module with other baseline fusion methods. Namely, element-wise addition and concatenation. They element-wise addition or concatenate the \textit{glimpse} vectors $\mathbf{g_t}$ and $\mathbf{g'_t}$ from the hybrid branch and the position enhancement branch before feeding them into the classifier at the $t^{th}$ time step during decoding. Different from dynamically-fusing, they are static and keep unchanged across different time steps and different decoding sequences.
Our proposed dynamically-fusing module achieves best results  on all 6 academical benchmarks and the second best result on RandText. Therefore, it is more robust than its static competitors, and applicable in more general application scenarios including those with or without context.

\begin{table*}{}
\setlength{\abovecaptionskip}{10pt}
\begin{minipage}{\linewidth}
\centering
\begin{tabular}{c|c|c|c|c|c|c|c}
\hline
 Method& IIIT5K & SVT & ICDAR 2013 & ICDAR 2015 & SVTP & CUTE 80 & RandText \\
\hline
Element-wise addition & 94.8 & 87.9 & 94.3 & 75.5 & 78.6 & 89.2 & \textbf{82.4} \\
Concatenation & 95 & \textbf{88.1} & 94.2 & 75.2 & 78.6 & 88.2 & 80.4 \\
Dynamically-fusing & \textbf{95.3} & \textbf{88.1} & \textbf{94.8} & \textbf{77.1} & \textbf{79.5} & \textbf{90.3} & 81.2 \\
\hline
\end{tabular}
\end{minipage}
\caption{Comparison with different fusion methods.}
\label{tab:df}
\end{table*}

\begin{table*}{}
\setlength{\abovecaptionskip}{10pt}
\centering
\begin{tabular}{c|c|c|c|c}
\hline
Method & SAR~\cite{li2018show} & DAN~\cite{wang2019decoupled}& Wang~\etal~\cite{wang2019simple}&RobustScanner\\ \hline
Accuracy & 29.9 & 51.1 & 46.4 & \textbf{55.7}\\ \hline
\end{tabular}
\caption{Comparison on the license plate dataset.}
\label{tab:cars}
\end{table*}

%\begin{figure}
%    \centerline{\includegraphics[width=0.9\linewidth]{res/images/lp.pdf}}
%    \caption{Samples of recognition results of our RobustScanner and SAR~\cite{li2018show} on Cars Dataset.}
%    \label{fig:lp}
%\end{figure}

\begin{figure}[h]
    \centerline{\includegraphics[width=0.9\linewidth]{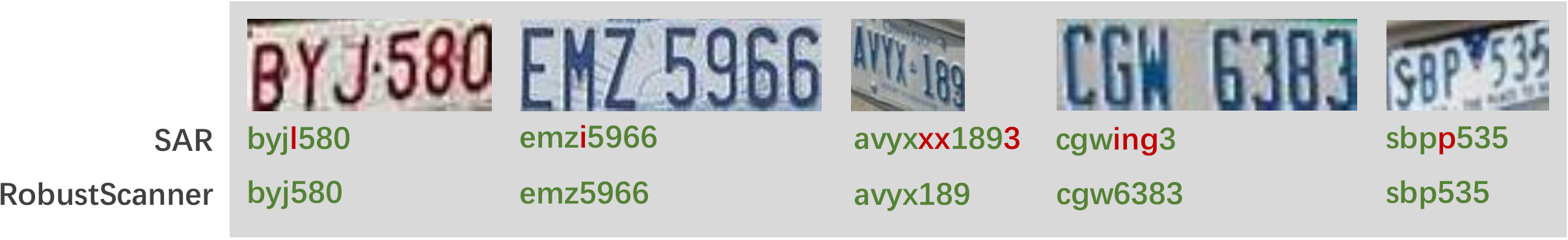}}
    \caption{Samples of recognition results of our RobustScanner and SAR~\cite{li2018show} on the license plate dataset.}
    \label{fig:lp}
\end{figure}

\textbf{Performance on license plate recognition.} To verify the effectiveness of our RobustScanner on contextless  data, we conduct experiments on the license plate recognition task.
We crop 97 license plate images from Cars dataset~\cite{KrauseStarkDengFei-Fei_3DRR2013} as the test set by using the character-level training annotations introduced in \cite{silva2018a}.
Noted that all methods are trained on the training data of MJSynth~\cite{jaderberg2014synthetic} and SynthText~\cite{gupta2016synthetic}. 
Therefore, the training and test data are very different in terms of background, font, and layout. All methods get relatively low accuracy on it. However, our RobustScanner outperforms other encoder-decoder with attention based methods with impressive margins, as shown in Table~\ref{tab:cars}.
Figure~\ref{fig:lp} shows some results of our method and SAR~\cite{li2018show} on the license plate test set.

\section{Conclusions}
In this paper, we target at mitigating the misrecognition problem of the encoder-decoder with attention framework on contextless text images. We have investigated the decoding procedure and
have found that the \textit{query} vectors of attention in the decoder contains both positional and context information, and positional clues become weak while contextual ones become strong as the decoding time step increases. Motivated by this finding, we have proposed RobustScanner for scene text recognition, which contains one hybrid branch and one position enhancement branch, and dynamically fuses the two branches at each time step during decoding. Moreover, we have proposed one novel position aware module, which can strengthen the positional encoding capacity of the position enhancement branch. 
We have conducted ablation studies to validate the effect of each proposed component.
We have extensively evaluated our proposed RobustScanner on both academical benchmarks and our synthesized RandText. The experimental results show that our approach has achieved new state-of-the-art results on popular regular and irregular text benchmarks while without much performance drop on contextless benchmarks. It has been validated that our proposed RobustScanner is robust in both 
context and contextless application scenarios.

\clearpage
\bibliographystyle{splncs04}
\bibliography{egbib}
\end{document}